\documentclass[conference]{IEEEtran}
\IEEEoverridecommandlockouts
\usepackage{cite}
\usepackage{amsmath,amssymb,amsfonts}
\usepackage{algorithmic}
\usepackage{graphicx}
\usepackage{textcomp}
\usepackage{xcolor}
\usepackage{booktabs}
\usepackage{subcaption}
\usepackage{listings}
\usepackage{bbding}
\usepackage{pifont}
\usepackage{wasysym}
\usepackage{amssymb}
\usepackage{outlines}
\usepackage{hyperref}
\usepackage{float}

\usepackage{array} %

\def\BibTeX{{\rm B\kern-.05em{\sc i\kern-.025em b}\kern-.08em
    T\kern-.1667em\lower.7ex\hbox{E}\kern-.125emX}}
\begin{document}
\title{A Multi-Agent Reinforcement Learning Testbed for Cognitive Radio Applications}

\author{
    \IEEEauthorblockN{
        Sriniketh Vangaru\IEEEauthorrefmark{1},
        Daniel Rosen\IEEEauthorrefmark{1},
        Dylan Green\IEEEauthorrefmark{1},
        Raphael Rodriguez\IEEEauthorrefmark{1},
        Maxwell Wiecek\IEEEauthorrefmark{1},\\
        Amos Johnson\IEEEauthorrefmark{2},
        Alyse M. Jones\IEEEauthorrefmark{1},
        William C. Headley\IEEEauthorrefmark{1}
    }
    \IEEEauthorblockA{
        \IEEEauthorrefmark{1}Virginia Tech National Security Institute,
        \IEEEauthorrefmark{2}Morehouse College\\
        amos.johnson@morehouse.edu, 
        alysemjones@vt.edu, 
        cheadley@vt.edu
    }
}

\maketitle

\bigskip
\section{Abstract}
\label{sec:abstract}
Technological trends show that Radio Frequency Reinforcement Learning (RFRL) will play a prominent role in the wireless communication systems of the future.  Applications of RFRL range from military communications jamming to enhancing WiFi networks.  Before deploying algorithms for these purposes, they must be trained in a simulation environment to ensure adequate performance. For this reason, we previously created the RFRL Gym: a standardized, accessible tool for the development and testing of reinforcement learning (RL) algorithms in the wireless communications space.  This environment leveraged the OpenAI Gym framework and featured customizable simulation scenarios within the RF spectrum.  However, the RFRL Gym was limited to training a single RL agent per simulation; this is not ideal, as most real-world RF scenarios will contain multiple intelligent agents in cooperative, competitive, or mixed settings, which is a natural consequence of spectrum congestion. Therefore, through integration with Ray RLlib, multi-agent reinforcement learning (MARL) functionality for training and assessment has been added to the RFRL Gym, making it even more of a robust tool for RF spectrum simulation. This paper provides an overview of the updated RFRL Gym environment. In this work, the general framework of the tool is described relative to comparable existing resources, highlighting the significant additions and refactoring we have applied to the Gym.  Afterward, results from testing various RF scenarios in the MARL environment and future additions are discussed.

\bigskip

\begin{IEEEkeywords}
multi-agent reinforcement learning, wireless communications, dynamic spectrum access, OpenAI Gym
\end{IEEEkeywords}

\section{Introduction}
\label{sec:intro}
\bigskip
\subsection{Radio Frequency Reinforcement Learning}
\label{sec:intro_rfrl}
In the wireless spectrum, the pre-allocation of specific frequencies by the FCC \cite{table_of_frequency_allocations} and the increasing usage of the radio frequency (RF) range of the spectrum, such as in mobile networks \cite{spectrum_allocation_accenture}, have motivated the use of cognitive radios (CR). CRs use dynamic spectrum access (DSA) to adaptively accommodate the quantity of signals continuously being sent and received. However, to enable these technologies to better manage the constantly fluctuating nature of the RF spectrum, Radio Frequency Reinforcement Learning (RFRL) technologies have been developed \cite{wireless_drl_overview}. These make prediction-based decisions regarding the frequencies a CR will transmit on as opposed to being purely reactive, and they have significantly improved the ability of a CR to avoid interference in the spectrum \cite{rl_based_channel_selection, rl_based_wideband, anti_jamming_using_rl}.

With the expansion of 5G and the early development of 6G wireless communication, demand for deployment-ready RFRL algorithms has increased dramatically; the RFRL Gym was created as a simulation environment and testbed to accelerate their development \cite{rfrl_gym}. However, the original version of the RFRL Gym included exclusively single-agent training scenarios. In the majority of real-world scenarios, multiple signals will exist in a wireless spectrum at one time, such as in military contexts where spectrum sharing and radio jamming are an essential concern \cite{5g_military} or in everyday products (e.g., mobile phones) whose ubiquity and usage of the spectrum are continuously increasing \cite{5g_spectrum_sharing}. CRs that act in the same space as other ``intelligent'' radios are dependent upon Multi-Agent Reinforcement Learning (MARL) methodologies to dependably achieve optimal results, as using a single-agent model will not properly encapsulate all of the dynamics of a given real-world situation; as a result, these are currently being rapidly researched and developed \cite{spectrum_sharing_drl, marl_allocation_framework, single_multi_drl_wireless}. Therefore, to increase the fidelity and realism of the training process, MARL scenarios were deemed a necessary addition to the testbed. This paper describes this primary addition to the RFRL Gym along with IQ data generation and rendering improvements for a more realistic, friendly user experience.

\subsection{Multi-Agent Reinforcement Learning}
\label{sec:intro_marl}
\begin{figure*}[hbt!]
\centering
    \includegraphics[width=\textwidth]{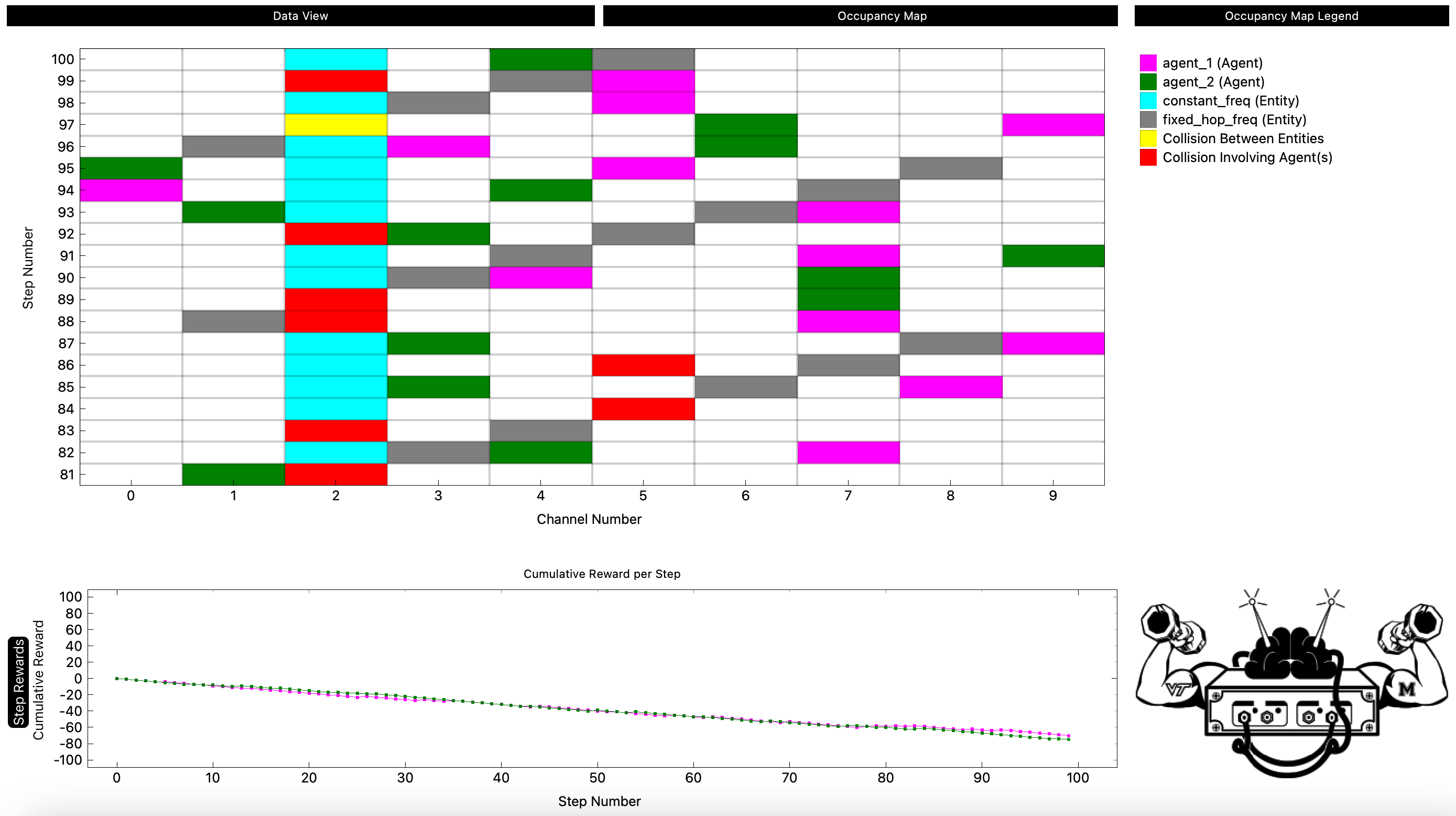}
    \caption{A rendering of an abstracted RF spectrum in the multi-agent RFRL Gym.}
    \label{fig:multi_agent_rendering}
\end{figure*}

Multi-Agent Reinforcement Learning (MARL) involves reinforcement learning scenarios in which multiple RL agents are acting and/or being trained simultaneously. In the context of the RFRL Gym, each agent is analogous to a CR device. Consider Fig. \ref{fig:multi_agent_rendering}, in which 4 distinct co-channel entities can be seen during the same time period.  Assume these signals are each being transmitted by an independent CR (a frequently-encountered scenario). In response to additional transient or persistent signals in the spectrum, the CRs may have to alter their transmission frequencies to avoid interference. Under the assumption that these CRs utilize RL to determine subsequent frequencies, this scenario is considered MARL. Such problems will become even more common as the expansion of IoT brings an increase in the number of devices that utilize the wireless spectrum, further emphasizing the need for programs such as our RFRL Gym to play a pivotal role in solving these problems.

In MARL, the reward functions of these agents are related to the actions of the other agents, since those actions would contribute to modifying the environment and make it non-stationary \cite{marl_selective_overiew}. Agents can be encouraged to act cooperatively, competitively, or a mixture of the two.  Independently from that categorization of situations, the MARL algorithms themselves can be centralized or decentralized. In a centralized context, a single model is used to determine the actions of multiple agents, whereas in a decentralized context, each agent is associated with its own model and determines its action exclusively.  Optionally, decentralized agents can communicate with each other in some manner---such as by sharing observations, joint rewards, or joint actions---which is a quality known as ``networked agents" \cite{marl_selective_overiew}. Our RFRL Gym is able to support all combinations of cooperative and competitive situations due to easy modification of scenarios, allowing for MARL research in RFRL to be further explored.

\subsection{Code Availability}
\label{sec:code_availability}

The code and data produced during the process of conducting this research are available at \url{https://github.com/vtnsi/rfrl-gym}.

\section{Background}
\label{sec:background}

\subsection{Related MARL}
\label{sec:background_marl}

Machine learning techniques have appeared in various facets of the RF Field \cite{cnns_deepsig, lstm_channel_simulation, ml_in_adv_rf}.  In particular, DSA has become a focus of RL research, both with singular agents \cite{deep_rl_dsa} and for multiple agents \cite{deep_marl_dsa}.  MARL methods have shown excellent performance in both cooperative \cite{marl_cooperative} and adversarial \cite{marl_adv} scenarios.  In recognition of these achievements, research efforts for MARL in RF have swelled \cite{marl_allocation_framework, single_multi_drl_wireless, marl_threshold_cr, marl_dsa_cr}.  Therefore, the addition of MARL functionality to the RFRL Gym was essential.

\subsection{Related Training Environments} 
\label{sec:background_envs}

\begin{table}[t]
    \caption{Comparison of existing OpenAI Gym tools for RL in wireless communications, modified from \cite{rfrl_gym}.}
    \begin{tabular}{p{0.1cm}p{1.7cm} p{0.6cm} p{1.0cm} p{1.1cm} p{0.5cm} p{1.0cm}p{0.1cm}}
    \toprule
    \vspace{4mm}
    \text{} &
    \textbf{Features} & 
    \textbf{RFRL Gym} &
    \textbf{GrGym \cite{gr_gym}} &
    \textbf{ns3-Gym \cite{ns3_gym}} &
    \textbf{Colosseum \cite{colosseum}} &
    \text{} \\
    \midrule

    \text{} &
    Flexible Scenario Design & \vspace{0.5mm} \hspace{1mm} \Checkmark & \vspace{0.5mm} & \vspace{0.5mm} & \hspace{1mm}  \vspace{0.5mm}\\
    
    \vspace{0.5mm} \\

    \text{} &
    RL Package Compatibility & \vspace{0.5mm} \hspace{1mm} \Checkmark & \vspace{0.5mm} \hspace{1mm} \Checkmark &  \vspace{0.5mm} \hspace{2mm} \Checkmark \\
    
    \vspace{0.5mm} \\
    
    \text{} &
    Spectrum Sensing Capabilities & \vspace{0.5mm} \hspace{1mm} \Checkmark & \vspace{0.5mm} & \vspace{0.5mm} \hspace{2mm} \Checkmark & \vspace{0.5mm} \hspace{1mm} \Checkmark \\
    
    \vspace{0.5mm} \\ 
    
    \text{} &
    Multi-Agent Capabilities & \vspace{0.5mm} \hspace{1mm} \Checkmark & \vspace{0.5mm} & \vspace{0.5mm} \hspace{2mm} \Checkmark & \vspace{0.5mm} \hspace{1mm} \Checkmark\\
    
    \vspace{0.5mm} \\
    
    \text{} &
    Ease of Use & \vspace{0.5mm} \hspace{1mm} \Checkmark & \vspace{0.5mm} & \vspace{0.5mm} & \vspace{0.5mm}\\
    
    \vspace{0.5mm} \\
    
    \text{} &
    Graphical User Interface & \vspace{0.5mm} \hspace{1mm} \Checkmark & \vspace{0.5mm} & \vspace{0.5mm} & \vspace{0.5mm} \hspace{1mm} \Checkmark \\
    
    \vspace{0.5mm} \\
    
    \text{} &
    Hardware Compatible & \vspace{0.5mm} & \vspace{0.5mm} \hspace{1mm} \Checkmark & \vspace{0.5mm} \hspace{2mm} \Checkmark & \vspace{0.5mm} \hspace{1mm}\\\bottomrule
    \end{tabular}
    \label{tab:feature_comparison}
\end{table}

Prior works \cite{gr_gym, ns3_gym, colosseum, rfrl_gym} describe tools for training RL algorithms for the wireless communications space, including the multi-agent RFRL Gym that we are introducing.  Table \ref{tab:feature_comparison} displays the differences between these environments.

\textbf{Flexible Scenario Design:} Ability to easily customize entities and agents present in the training environment.  \cite{gr_gym} scenarios can be customized, but each requires a non-trivial implementation of a scenario class for each testing scenario.  Similarly, \cite{ns3_gym} requires a custom C\(++\) interface to be written for each scenario. \cite{colosseum} provides a wide variety of premade scenarios of incumbents (similar to entities).  Though not customizable, they are relatively comprehensive in conjunction.  The RFRL Gym's scenarios can be easily and finely customized with JSON scenario configuration files.

\textbf{RL Package Compatibility:} Ability to interface with external RL libraries for algorithms and optimization methods.  \cite{gr_gym, ns3_gym}, and the RFRL Gym utilize OpenAI's Gym API \cite{openai_gym}.  Because of this, these environments interface with various libraries including Stable-Baselines \cite{sb3}, PettingZoo \cite{petting_zoo}, etc.  \cite{colosseum} is not similarly integrated, as models are primarily intended to be trained offline on data collected from the environment.

\textbf{Spectrum Sensing Capabilities:} Various methods of observing the state of the spectrum.  This includes methods of detecting and classifying entities. 

\textbf{Multi-Agent Capability:} The newest version of the RFRL Gym environment features the addition of multi-agent capabilities, meaning that it can support the concurrent training of multiple intelligent agents in one scenario. This also includes centralized and decentralized information structures, as well as competitive, cooperative, and mixed settings for scenarios.

\textbf{Ease of Use:} Tool is designed to simplify the training and analysis process of RL algorithms, especially if this functionality is accessible to a wide audience.  \cite{gr_gym} and \cite{ns3_gym} both require understanding of an external resource (GNU Radio and ns-3 respectively).  GNU Radio is not officially supported on Windows and \cite{gr_gym} does not offer rendering.  \cite{ns3_gym} requires C\(++\) implementation to interface with the ns-3 environment.  \cite{colosseum} does not offer rendering and access is limited to researchers affiliated with the US Department of Defense.  The RFRL Gym is designed to be accessible to both those new to wireless communications and those with extensive experience in the field.  The RFRL Gym features a GUI for easy scenario generation and a variety of rendering options for data visualization.

\textbf{Hardware Compatibility:} The training environment's ability to interface with radio hardware for real-time information collection. \cite{gr_gym} and \cite{ns3_gym} both utilize external resources for wireless spectrum simulation, GNU Radio and ns-3 respectively.  Both of these resources can interface with hardware for real-time spectrum sensing or simulate the wireless spectrum with software.  \cite{colosseum} uses fiber optics to emulate the wireless spectrum.  So, although its resources are accessed remotely by users, no real radios are involved.  However, because 
\cite{colosseum} adopts the OpenRAN Gym framework \cite{open_ran_gym}, containerized applications can easily be ported from Colosseum to hardware-interfacing training environments.  Currently, all of the RFRL Gym's spectrum simulation is software-driven using open-source Python modules but the addition of hardware compatibility is in progress.
 
Relative to comparable tools, the RFRL Gym offers a uniquely accessible workflow.  The environment features deeply customizable scenarios for bespoke training executions.  With human-readable information from these scenario files, IQ data is generated using abstracted information so a limited understanding of wireless communications is necessary for users. These scenario files can be produced using an easy-to-use graphical user interface, as shown in \cite{rfrl_gym}.  Overall, without the burden of the need for extensive knowledge of wireless communications, researchers can use this tool to incrementally develop an understanding of concepts such as spectrum sensing, DSA, and signal modulation.  The robust visualization and efficient abstraction make the RFRL Gym an effective educational tool for those new to the field of RF.  These features distinguish our tool from others as an accessible resource for experimentation and testing.  However, it shares its more technical features and functionality with powerful spectrum simulation tools, such as GNU Radio and ns-3, making the RFRL Gym an appropriate tool for experts as well.

\section{The RFRL GYM MARL Framework}
\label{sec:framework}

\begin{figure*}[h] 
    \begin{center}
        \includegraphics[width = \textwidth]{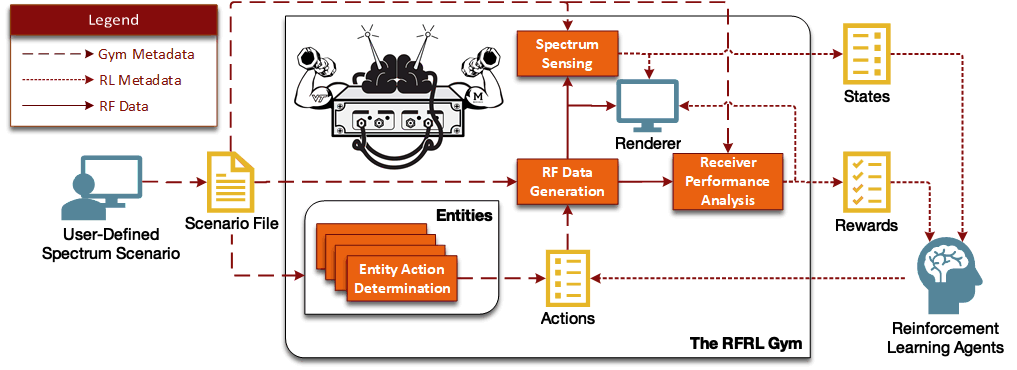}
    \end{center}
    \caption{Diagram of information flow in the multi-agent RFRL Gym.}
    \label{fig:flow_diagram}
\end{figure*}

\subsection{Workflow}
\label{sec:framework_workflow}

In order to execute training simulations, users will first create a JSON scenario file.  This file contains information about the agents and environment (detailed below in Section \ref{sec:framework_scenario}).  Fig. \ref{fig:flow_diagram} shows the exchange of information that occurs during a training simulation.  Entities and agents are prompted by the gym to select actions at each timestep; entities follow the action pattern corresponding to the entity type provided in the scenario, and learning agents take actions based on their policies that were refined through a learning algorithm.  This data is optionally converted into IQ data if using the single-agent environment. Then, the environment determines the occupancy of each channel and displays its results using the selected rendering method. Investigators can use this displayed information to assess the performance of their algorithms. Afterward, the agents' rewards and the environment's state are calculated, and these are recorded by the gym before being sent to the agents to begin the next timestep.

\subsection{Comparison to Original RFRL Gym}
\label{sec:framework_comparison}
This paper expands upon the existing single-agent version of the RFRL Gym, defined in \cite{rfrl_gym}.  Structurally, the new version of the RFRL Gym remains similar to that of the initial version.  As can be seen in Fig. \ref{fig:flow_diagram}, information transfer between components remains consistent in the MARL environment, abiding by the classic reinforcement learning cycle.

The original single-agent environment will still exist in the package with its previous functionality. One addition made to this environment is a separate single-agent environment class that incorporates software-generated IQ data to better consider real spectrum dynamics, described further in Section \ref{sec:framework_iq}. Another primary modification made to the original environment is the upgrade to the Farama Foundation's Gymnasium \cite{gymnasium}, which is an extension to OpenAI's Gym API \cite{openai_gym}.  This interface is compatible with external RL libraries such as Stable-Baselines3 \cite{sb3} and PettingZoo \cite{petting_zoo}. This API is also leveraged by our MARL environment through inheritance from Ray RLlib's \texttt{MultiAgentEnv} class \cite{rllib_documentation_pages}. On top of compatibility with external libraries, Ray RLlib provides several high-powered functions such as remote cluster computation and cloud interfacing, whose computing power can be helpful when running larger-scale MARL simulations.

In the new multi-agent environment, the movement patterns that the non-learning RF entities have available in the RL environment have not been modified. Their functions and applications can be found in \cite{rfrl_gym}. Each agent can track the locations of the entities with either the \texttt{detect} or \texttt{classify} observation mode, as in the original Gym.  The observation and action histories of each agent are stored independently and provided to the user at each timestep, facilitating easy analysis of individual agents.  The primary differences between the MARL RFRL Gym environment and the single-agent RFRL Gym environment exist in reward calculation, scenario file construction, data structures for observation and action histories, and rendering performance.  The rest of this section enumerates and explains these features in further detail.

\subsection{MARL Reward Calculation}
\label{sec:framework_reward}
Ray RLlib's RL training framework includes user-defined agent groups, which can, for instance, be defined explicitly using RLlib's \texttt{MultiAgentEnv.with\_agent\_groups()} function or implicitly using the \texttt{policy\_mapping\_fn} configuration property depending on the algorithm being used. The members of an agent group can view fellow members' actions at each timestep. The rewards for each agent in a group are correlated, and so these features permit centralized RL algorithms to be trained in our environment. This is useful as many algorithms use a very common approach involving centralized training with decentralized execution (CTDE) \cite{centralized_vs_decentralized_critics}. A centralized critic can even be implemented manually, such as by creating a function that shares observations \cite{rllib_documentation_pages}.

Alternatively, agent groupings with one member per group may simulate a decentralized reward system. A standard use case for this method is decentralized training with decentralized execution (DTDE), a step up from independent Q-learning (IQL) in a range of scenarios, including mixed cooperative-competitive settings \cite{NING2024, dtde}.

The individual reward functions are given in Equations 1 and 2, where \textit{a} is an agent's action and \textit{g} is the set of actions from the agents in a group. The dynamic spectrum access (DSA) function rewards the agent for avoiding channels containing other entities or agents, and the jamming function rewards the agent for occupying the same channel as a particular entity in the environment, which is selected by the user. The reward function for each agent group is calculated as the sum of the individual rewards of each agent \cite{rllib_documentation_pages}. The mean reward, attributed to each agent in an agent group, can be accessed in the \texttt{results} dictionary returned by RLlib during training but its usage (compared to the usage of the group's reward) depends on the algorithm used. This is shown in Equation 3. As a result of this customizability of agent groups, the RFRL Gym is suitable for cooperative, competitive, and mixed training scenarios.

\begin{equation}
    r_{DSA}(a) =
    \begin{cases}
        1, &  \text{no collision} \\
        0, & \text{no transmission} \\
        -1, &  \text{collision}
    \end{cases}
\end{equation}
\begin{equation}
    r_{jamming}(a) = 
    \begin{cases}
        1, &  \text{collision with target entity} \\
        0, & \text{no transmission} \\
        -1, &  \text{transmitting elsewhere}
    \end{cases}
\end{equation}
\begin{equation}
r_{mean}(g) = \frac{\sum_{a \epsilon g}{r(a)}}{{|g|}}
\end{equation}

It is important to note that the amount of centralization that is involved in training depends mainly on the implementations of the algorithms themselves and not the RFRL Gym environment\footnote{Customized reward functions in the environment as mentioned in Section \ref{sec:conc} could also influence centralization, however.}, and the agent groupings involved in the testing scripts are made to correspond with the form of centralization used. Additionally, a decentralized setup is the most common for practical scenarios with CRs in the RF spectrum, as centralization would imply a control unit that connects independent CR devices in the real world, which is not usually feasible. As a result, we only used the decentralized option of agent groups for our evaluations and did not conduct further testing with different groupings or reward function centralization. Though our testing scripts do still illustrate the use of RLlib's representation of groups/centralization (as our decentralized setup is just a special case of agent groups but with one agent per group), that was not the main focus of our demonstration of the RFRL Gym environment.

\subsection{MARL Scenario Files}
\label{sec:framework_scenario}
The definition of a scenario and its separate categories are stated in \cite{rfrl_gym}. The only category that has changed is the \texttt{environment} dictionary in the JSON file, which controls all specific constraints that the multiple agents will learn under. 
Whereas our previous implementation only included the configurations for one agent, the \texttt{environment} section now contains an \texttt{agents} dictionary, which defines the number of agents in the environment and the observation mode, reward function, and target entity (if applicable) for each agent. For ease of use, we have also added a \texttt{comments} key-value pair to allow the user to add a description of the scenario within the JSON itself. This ensures that when the scenario is given to another analyst, it helps them understand the scenario's contents and purpose.

\subsection{Rendering Enhancement}
\label{sec:framework_rendering}
Previous rendering in the RFRL Gym only supported a single-agent environment for both PyQt rendering and terminal rendering, as explained in the previous paper \cite{rfrl_gym}. With the enhancement of the RFRL Gym to a multi-agent environment, the major changes in rendering include visuals of cumulative reward changes for each agent at each time step in a training episode (or at each episode, depending on the user's configuration) and different colors to represent each learning agent or non-learning entity as it occupies a space in the radio frequency spectrum, as shown in Fig. \ref{fig:multi_agent_rendering}. The same color that identifies a specific agent in the depiction of the spectrum will also be used for the corresponding agent on the cumulative reward graphs. The primary justification behind giving a different color scheme to each individual agent, similarly to how multiple entities in the single-agent gym would have their own color, is to easily identify which agents are learning their action policy effectively. The subclasses of our \texttt{MultiAgentRenderer} class can be modified to visually distinguish between different types of collisions (e.g., agent with agent, agent with entity, entity with entity) by changing the colors that represent such events.

\subsection{IQ Data Generation}
\label{sec:framework_iq}
The RFRL Gym also now includes another single-agent environment in the form of a Python class (\texttt{RFRLGymIQEnv}) that uses software-generated IQ data to represent the locations of entities and the agent in the spectrum.  In this environment, at each time step, the discrete actions selected by each entity and the agent are used to produce simulated IQ data.  Modulation methods are available to users out of the box in both \texttt{LDAPM} and \texttt{Tone} classes. Once the data are created, the environment infers occupancy of each channel based on an energy threshold being met.  This method improves the realism of scenarios as the non-deterministic nature of the wireless spectrum is included in the training.

\section{Simulation Results and Discussion}
\label{sec:results}

\subsection{Algorithms}
\label{sec:results_algorithms}

Four multi-agent reinforcement learning algorithms were investigated in our evaluations. They were chosen due to being RLlib's only available algorithms with multi-agent functionality at the time of testing. For each, the \texttt{policy\_mapping\_fn} function in the algorithm's configuration maps each agent to its own policy so that, as mentioned earlier in Section \ref{sec:framework_reward}, the agent groups simulate a decentralized setting. The algorithms are listed below.

\subsubsection{Deep Q-Network (DQN)} DQN is a widely-used RL algorithm that uses a neural network to approximate Q-Values \cite{dqn}. The DQN implementation used for simulations uses an experience replay buffer to minimize recency bias, which prioritizes selecting more recent experiences. Though DQN is originally a single-agent learning algorithm, RLlib's multi-agent extension of DQN uses a prioritized replay buffer, designed specially for multi-agent environments, that allows the user to optimize how the prior sampled experiences are picked. DQN is typically well-suited for discrete action spaces \cite{action_space_drl}.

\subsubsection{Proximal Policy Optimization (PPO)} PPO is another well-known, standard learning algorithm in single-agent RL systems, using an objective function and a surrogate objective function where the policy updates for the latter are constrained to remain relatively small \cite{ppo}. This tends to ensure that learning is stable. Similarly to DQN, we used RLlib's implementation of multi-agent PPO.

\subsubsection{Asynchronous Proximal Policy Optimization (APPO)} APPO is an RLlib-implemented variant of PPO with asynchronous sampling of experiences, which allows it to be quicker (measured in real-world time) than PPO, though its performance is not always better than standard PPO \cite{rllib_documentation_pages}.

\subsubsection{Importance Weighted Actor-Learner Architecture (IMPALA)} IMPALA is an algorithm that specializes in multi-task learning but can also handle multi-agent environments \cite{impala}. It is based on the ``asynchronous advantage actor-critic'' (A3C) algorithm and in fact provides a similar architecture to what APPO uses.

\subsection{Evaluation Methods}
\label{sec:results_methods}

In order to test our scenarios, we wrote a testing script for each algorithm to run and evaluate the effectiveness of the learned policies for the agents in any given scenario. Due to each algorithm containing a different number of episodes per training iteration, the method of algorithm evaluation is based on training episodes to maintain uniformity. The collective reward accrued by the agents in an episode will be plotted against the episode count for each algorithm, with the results of the algorithms' performances shown and discussed below in Section \ref{sec:results_scenarios}.

\subsection{Scenarios Tested}
\label{sec:results_scenarios}

Here, we present a variety of scenarios which represent different states that learning agents and non-learning entities can take on in wireless communications situations, before discussing the results from training and evaluating the learning algorithms for the agents. For each scenario, the environment assumes a standard of 10 discrete frequency channels, as well as 100 sampled timesteps (i.e., calls of \texttt{env.step()}, where \texttt{env} is an instance of our multi-agent environment class) per episode of training. It should be noted that during training, depending on the algorithm used, the number of times that each sampled \texttt{step()} call is actually trained on varies and, as mentioned earlier, the number of episodes per iteration also varies. Additionally, as mentioned in Section \ref{sec:framework_reward}, we use a fully decentralized system for every scenario, so for each algorithm, each agent is mapped to its own policy.

\subsubsection{Scenario 1: Jamming Constant-Frequency Entities}
As a baseline test of the ability of our multi-agent RFRL Gym environment to support learning algorithms that train the agents' policies, a simple scenario was created where each agent was given an entity to jam (i.e. transmit on the same channel as). There are four agents, where two have the \texttt{detect} observation mode and the other two use the \texttt{classify}, and all of them are using the \texttt{jam} reward mode. There are also 2 entities which constantly remain on one frequency channel each, separate from each other.

\begin{figure}
    \includegraphics[scale = 0.4]{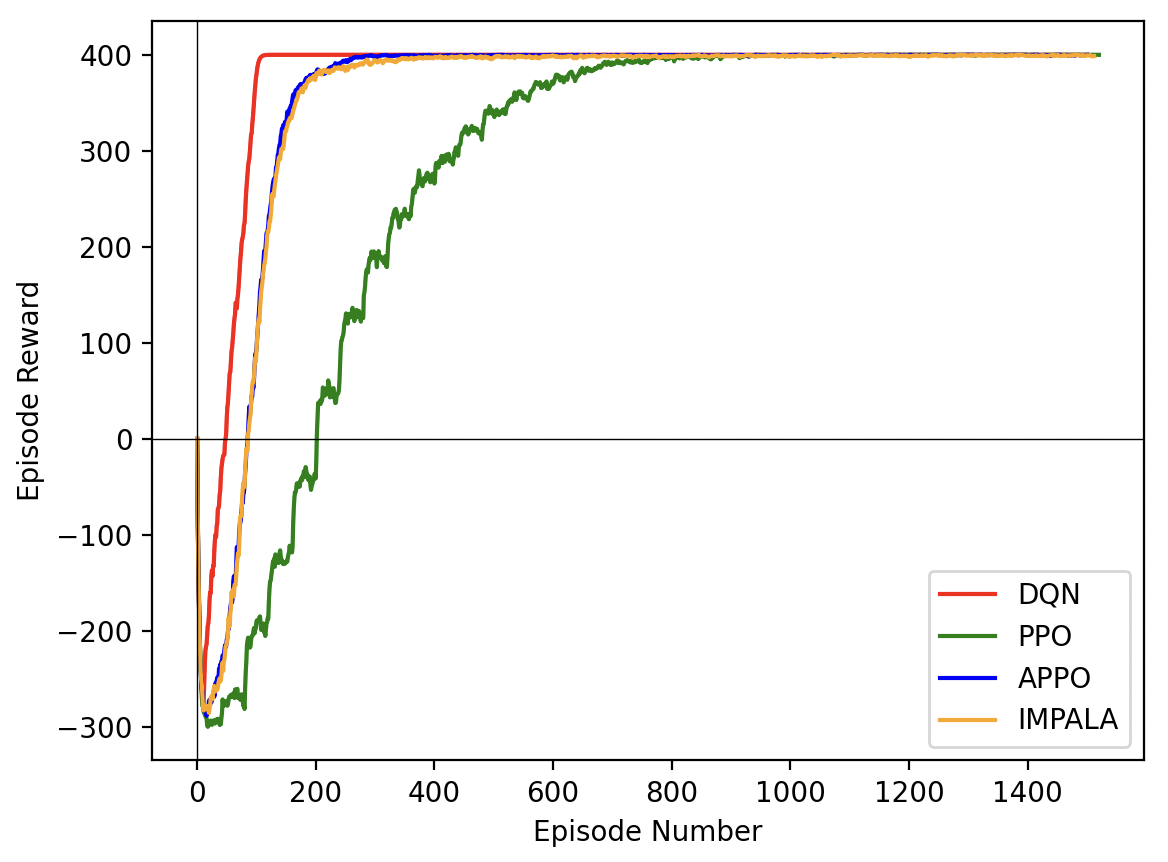}
    \caption{Results from testing Scenario 1. To lessen the impacts of episode reward fluctuations upon the display, an exponentially weighted moving average (EWMA) with $\beta=0.75$ was applied.}
    \label{fig:scenario_1}
\end{figure}

The environment successfully interfaced with the various aforementioned algorithms implemented by RLlib, and the results are displayed in Fig. \ref{fig:scenario_1}. An episode reward of 400 indicates that all 4 agents consistently picked the optimal channel to transmit on; this is because each optimal action obtains a reward of 1 for that agent, and the rewards are simply summed up across all 4 agents and 100 timesteps in an episode to form the episode reward. \textit{(Note: For the same reason, Scenario 5 also has a maximum episode reward of 400.)} As shown, all algorithms converged to an optimal policy for all agents, demonstrating the basic capabilities of our RFRL Gym environment involving having multiple agents effectively learn to complete a task in the RF spectrum. Evidently, DQN demonstrated faster convergence than the other algorithms, which is likely due to its general aptitude for discrete action spaces, with the benefit from this being compounded by only having a small number of actions---which are the 10 frequency channels to transmit on---available for each agent \cite{action_space_drl}.

\subsubsection{Scenario 2: Fixed-Hop Frequency Jamming}
\begin{figure}
    \includegraphics[scale = 0.4]{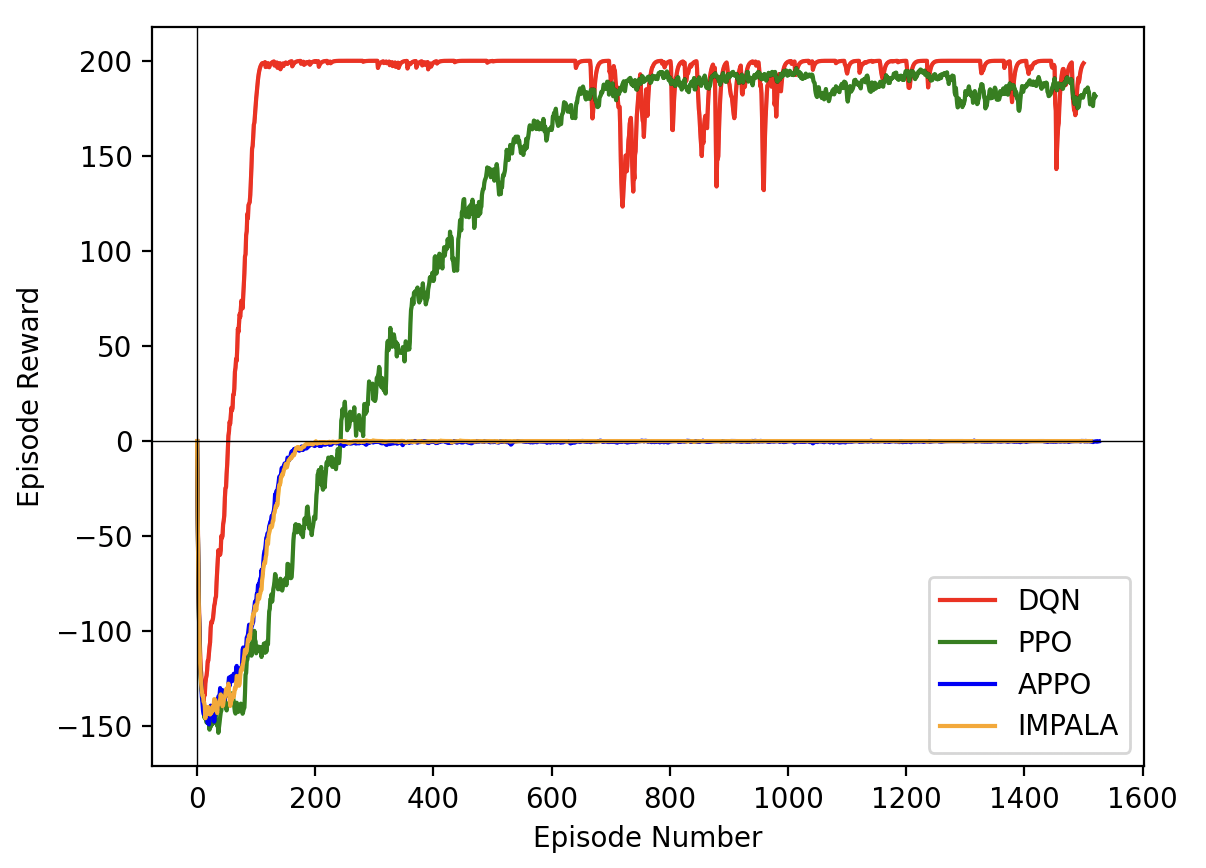}
    \caption{Results from testing Scenario 2. An EWMA with $\beta=0.75$ was used.}
    \label{fig:scenario_2}
\end{figure}

To incorporate slightly more realistic non-learning, transmitting entities compared to those in Scenario 1, we created a scenario which involves two agents trying to both jam a single entity that uses a linear hopping pattern. The results are shown in Fig. \ref{fig:scenario_2}. Due to only having 2 agents, the maximum episode reward attainable in this scenario is 200. \textit{Note: Due to the same reason, Scenarios 3 and 4 also have a maximum episode reward of 200.}

This scenario revealed more interesting results than Scenario 1, with the most noticeable difference being that APPO and IMPALA converged to an episode reward value of 0. After observing the agents' actions, it became clear that the two agents attempting to follow the fixed hopper entities were not able to learn the optimal actions to take---regardless of their observation mode---and instead opted to not transmit on any channel, giving a reward of 0, instead of repeatedly picking the wrong channel, which would give a reward of -1. A probable reason for this, considering that APPO and IMPALA share the same asynchronous sampling architecture for their replay buffer of experiences, is that the nature of the fixed-hopping entity requires the learning algorithm to be aware of the hopping pattern (which changed the transmission channel every step in time), and the asynchronous data collection in these algorithms may not be capable of that \cite{impala}.

Additionally, DQN demonstrated rather quick learning to an optimal convergence, before dropping its performance at around 600 episodes; this drop in performance is conventionally known as catastrophic forgetting \cite{catastrophic_forgetting}. Upon further investigation, a highly likely cause of this is that 600 episodes, or 60,000 steps, is equal to the size of the experience replay buffer capacity of this DQN implementation which was set as a hyperparameter, so the drop in episode rewards coincides with the point at which the replay buffer begins to filter out previous experiences, whether in a first-in first-out (FIFO) cyclic manner or with a different methodology \cite{dqn}. Allowing previous experiences to be removed from the sampling process can destabilize learning due to, for instance, specific valuable experiences being lost or the policy being learned from new experiences differing from the policy attained from the initial 60,000 experiences \cite{catastrophic_forgetting_2}. DQN did appear to relearn after this, but in this scenario, PPO demonstrated more robustness against catastrophic forgetting. Though this is specific to the family of algorithms that uses replay buffers, researchers using our tool to develop MARL algorithms might mitigate against errors like this by sampling different hyperparameter combinations or using different buffer structures, such as hybrid buffers that permanently store select few experiences or separately store rare transitions.

\subsubsection{Scenario 3: Dynamic Spectrum Access with Constant Frequency Entities}
\begin{figure}
    \includegraphics[scale = 0.4]{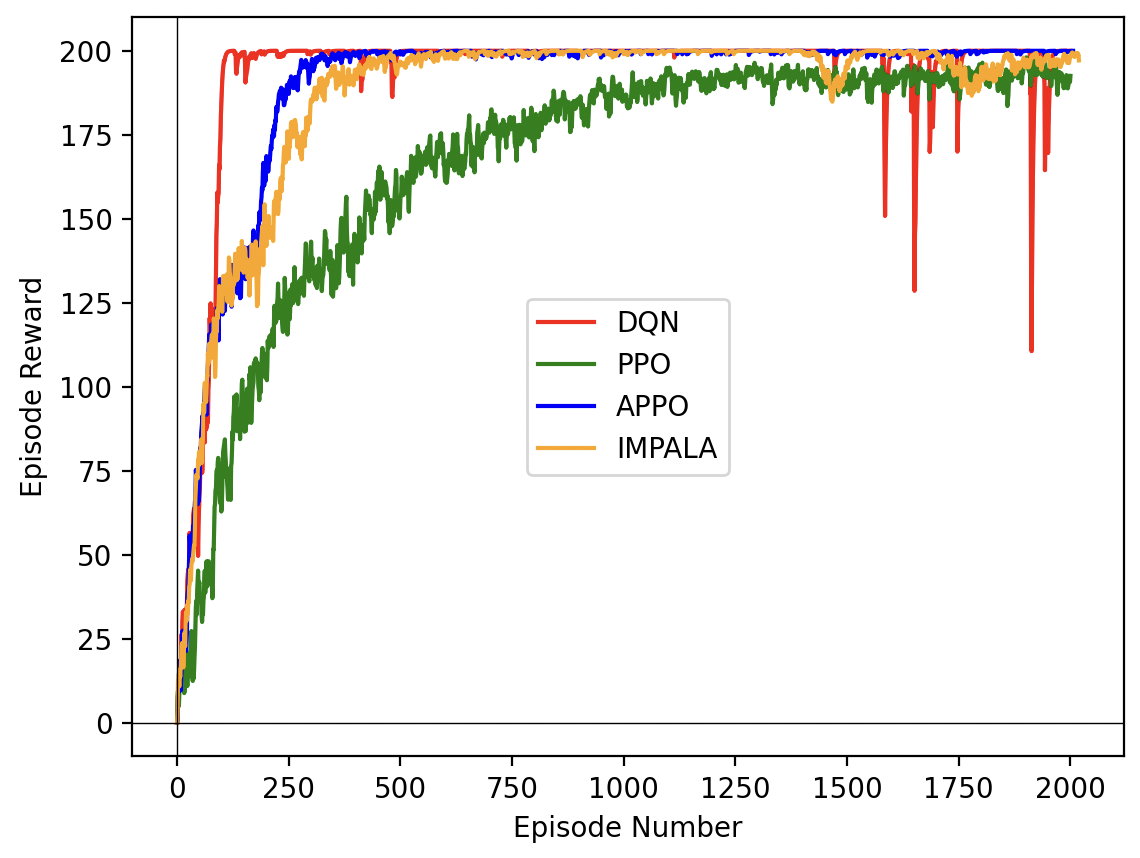}
    \caption{Results from testing Scenario 3. An EWMA with $\beta=0.75$ was used.}
    \label{fig:scenario_3}
\end{figure}

An extremely common setting in wireless communications is when each cognitive radio possesses the goal of transmitting on a free channel. This scenario was designed to simulate how RF devices would learn to act in a simple variation of this setting, by using the \texttt{dsa} reward function. Due to the non-learning entities only transmitting on constant frequencies in this scenario, it is comparable to Scenario 1 when each agent learned to target one set channel, except in this case multiple agents targeting the same channel would negatively reward both of them. It was expected that learning would be fairly fast, similarly to Scenario 1, but with less stability due to agents being able to interfere with each other's learning goals, and these predictions were reflected in the results. They are displayed in Fig. \ref{fig:scenario_3}. One notable outcome was that DQN once again demonstrated catastrophic forgetting after maintaining high rewards consistently.

\subsubsection{Scenario 4: Dynamic Spectrum Access with Fixed-Hop Frequency Entities} 
\begin{figure}
    \includegraphics[scale = 0.4]{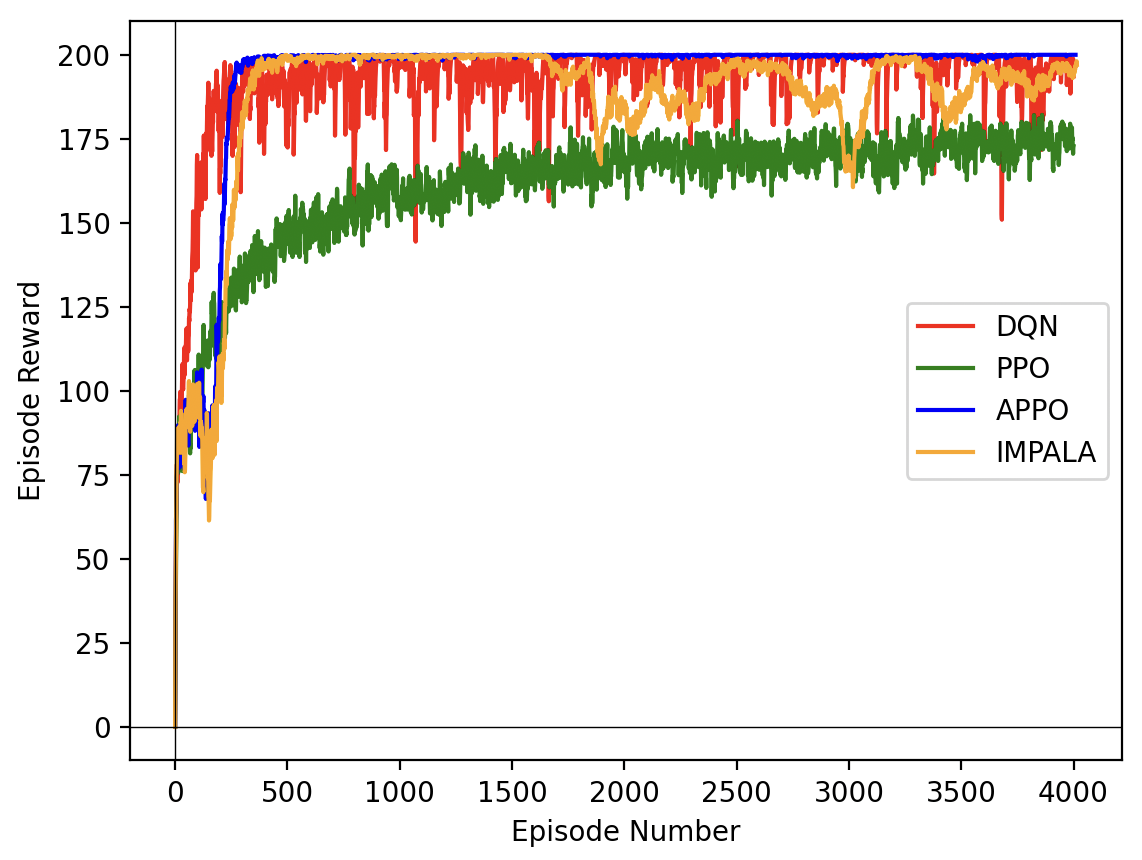}
    \caption{Results from testing Scenario 4. An EWMA with $\beta=0.75$ was used.}
    \label{fig:scenario_4}
\end{figure}

In this scenario, we wished to demonstrate agents learning to find free channels when there are moving entities in the environment. However, there are fewer entities than in the previous scenario to make it more feasible for the agents to identify patterns in the channels that are occupied. Naturally, there was significantly more variability in the rewards due to the agents having to learn a more complicated policy and continuously unintentionally running into collisions (e.g., with each other) in the convoluted spectrum; the results are in Fig. \ref{fig:scenario_4}.

\subsubsection{Scenario 5: A Mixed Setting} 
\begin{figure}
    \includegraphics[scale = 0.4]{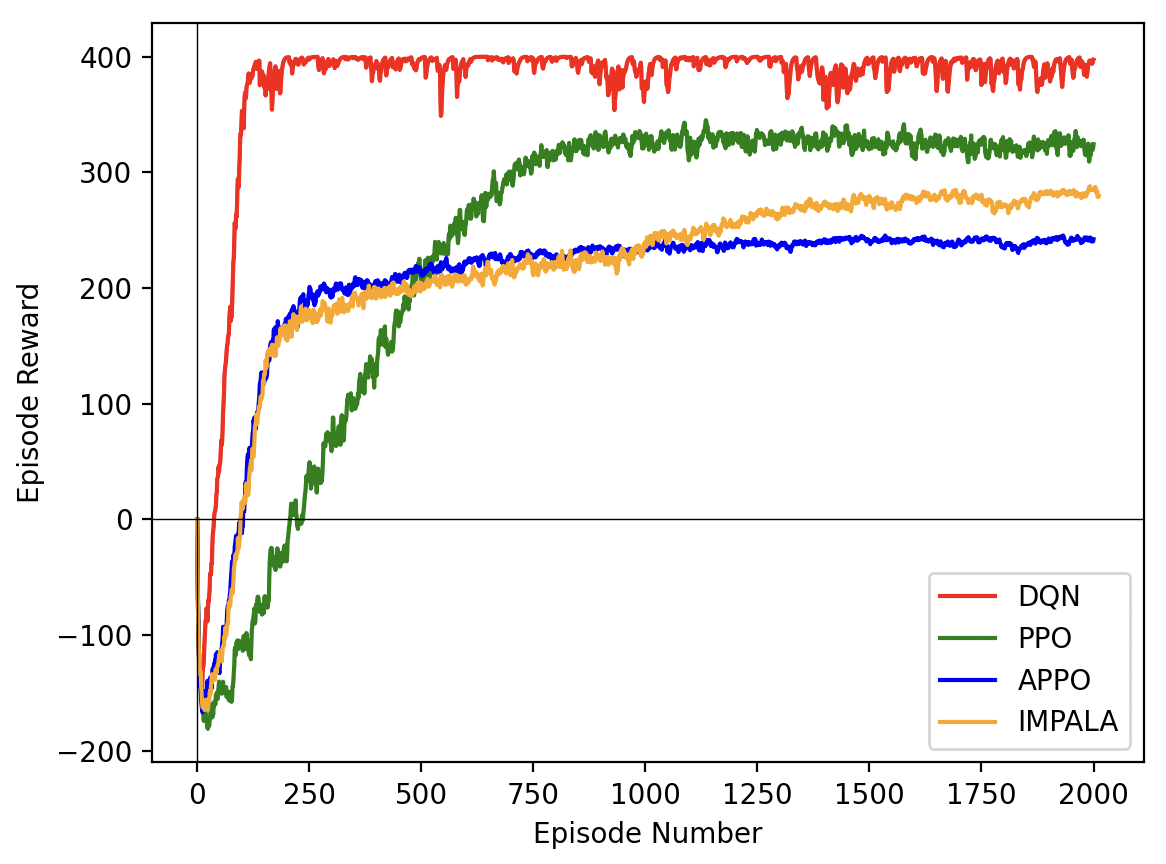}
    \caption{Results from testing Scenario 5. An EWMA with $\beta=0.75$ was used.}
    \label{fig:scenario_5}
\end{figure}

With each scenario so far, we have attempted to incorporate more realism to provide simulations of these algorithms working in useful contexts, so Scenario 5 emulates a situation with some agents having goals that allow for cooperation with others being directly adversarial to other agents in the spectrum. To support this, there is variance in observation modes, reward modes, and target entities among the 4 agents. The results are shown in Fig. \ref{fig:scenario_5}.

Evidently, none of the algorithms learned to approximate a completely optimal function. DQN learned an almost optimal policy, although there are continuous dips in performance which were due to the DSA agents taking actions that interfered with the fixed-hop frequency entities. Similarly, with each of PPO, APPO, and IMPALA, a subset of the agents learned a policy without issue and the remaining agents either decided to not transmit to obtain the 0 reward or continuously chose suboptimal actions that averaged out to a positive reward. Similar to the previous scenarios for these algorithms, the agents jamming the constant-frequency entities tended to learn perfectly, the agents jamming the fixed-hop frequency entities tended to not transmit due to difficulty following the entity, and the agents finding free channels did transmit but had high variability in their rewards. A primary goal in this tool's development is to empower future RF and RL researchers who are using it to explore exactly these types of nuanced trade-offs, as understanding them is crucial for developing future algorithms that reach optimal policies in situations with competing intelligences. Additionally, as these scenarios have shown that each algorithm has cases where it falters, these results further demonstrate the gap that can be filled by using our tool to construct sophisticated MARL algorithms targeted toward a variety of situations in radio communications.

\section{Conclusion}
\label{sec:conc}

This paper introduces the addition of multi-agent reinforcement learning functionality to the RFRL Gym.  This new functionality is key for accurately representing and simulating the real-world wireless spectrum, as evidenced by the increase in spectrum usage by various modern devices in recent years \cite{spectrum_allocation_accenture}.  In conjunction with this upgrade, the RFRL Gym was integrated with Ray RLlib.  This new interface leverages the new industry-standard Gymnasium API.  As a result, the RFRL Gym now supports countless additional integrations including Stable-Baselines3, PettingZoo, and others.

From experimentation, our results conclusively demonstrate that the environment properly enables multiple agents to learn optimal policies, despite having complex variations across the reward functions used. Additionally, it has been shown that a variety of multi-agent learning algorithms can be used in tandem with the new multi-agent RFRL Gym environment.

Future enhancements to the radio frequency (RF) modeling aspect of our testbed include the use of real hardware to integrate with and represent the state of the environment, adding multi-channel agents (which can represent real-world devices that transmit signals on multiple frequency bands at once), and implementing more complex channel modeling approaches to take into consideration more RF details when simulating multiple agents. On the machine learning front, future work to be done involves allowing for more customizable reward functions (e.g., functions that incentivize agents to jam other learning agents or that define collaboration strategies where only a single agent in the collaborating group jams an entity) within the scenario file, evaluating the performance (e.g., convergence time) of different algorithms in our testbed as the number of agents is scaled up (which, with the RLlib API, can be scaled up arbitrarily), and testing different variations of scenarios with full centralization / partial centralization of the agents' observations, actions, and rewards.

\bibliographystyle{IEEEtran}
\bibliography{references.bib}
\end{document}